\documentclass[10pt,twocolumn,letterpaper]{article}

\usepackage{wacv}
\usepackage{times}
\usepackage{epsfig}
\usepackage{graphicx}
\usepackage{amsmath}
\usepackage{amssymb}
\usepackage{multirow}

\def\wacvPaperID{715} 

\wacvfinalcopy 

\ifwacvfinal
\def\assignedStartPage{1} 
\fi

\ifwacvfinal
\usepackage[breaklinks=true,bookmarks=false]{hyperref}
\else
\usepackage[pagebackref=true,breaklinks=true,colorlinks,bookmarks=false]{hyperref}
\fi

\ifwacvfinal
\else
\fi

\newcommand{\bgamma}{{\mathbf{\gamma}}}
\newcommand{\REAL}{\ensuremath{\mathbb{R}}}
\newcommand{\stylegan}{\textit{StyleGAN}}
\newcommand{\stylerig}{\textit{StyleRig}}

\usepackage{cuted}
\usepackage{capt-of}
\begin{document}

\title{Unsupervised High-Fidelity Facial Texture Generation and Reconstruction}

\author{Ron Slossberg$^*$\\
Technion\\
{\tt\small ronslos@gmail.com}
\and
Ibrahim Jubran\thanks{Authors contributed equally.}\\
University of Haifa\\
{\tt\small ibrahim.jub@gmail.com}
\and 
Ron Kimmel\\
Technion\\
{\tt\small ron@cs.technion.ac.il}
}

\maketitle


\begin{abstract}
Many methods have been proposed over the years to tackle the task of facial 3D geometry and texture recovery from a single image.
Such methods often fail to provide high-fidelity texture without relying on 3D facial scans during training.
In contrast, the complementary task of 3D facial generation has not received as much attention.
As opposed to the 2D texture domain, where GANs have proven to produce highly realistic facial images, the more challenging 3D geometry domain has not yet caught up to the same levels of realism and diversity.

In this paper, we propose a novel unified pipeline for both tasks, generation of both geometry and texture, and recovery of high-fidelity texture. Our texture model is learned, in an unsupervised fashion, from natural images as opposed to scanned texture maps.
To the best of our knowledge, this is the first such unified framework independent of scanned textures.

Our novel training pipeline incorporates a pre-trained 2D facial generator coupled with a deep feature manipulation methodology.
By applying precise 3DMM fitting, we can seamlessly integrate our modeled textures into synthetically generated background images forming a realistic composition of our textured model with background, hair, teeth, and body.
This enables us to apply transfer learning from the domain of 2D image generation, thus, benefiting greatly from the impressive results obtained in this domain.

We provide a comprehensive study on several recent methods comparing our model in generation and reconstruction tasks.
As the extensive qualitative, as well as quantitative analysis, demonstrate, we achieve state-of-the-art results for both tasks.
\end{abstract}


\section{Introduction}

Generation of 3D facial geometry and full texture, as well as their reconstruction from a single 2D image, are highly challenging and important tasks at the intersection of computer vision, graphics, and machine learning. 
These tasks arise within endless applications ranging from virtual reality and computer gaming to facial editing.

At the heart of such generation and reconstruction methods lies a hidden common assumption that natural facial geometries and textures reside on a low-dimensional manifold.
Following this assumption, the above tasks can be carried out within this simpler representation space, instead of the original high-dimensional space. 
The recovery of this manifold is termed \emph{facial modeling} and the mathematical bridge between the high and low dimensional representations is termed a \emph{facial model}.

Many different types of facial models have been in use, including linear models, non-linear deep learning-based models, hybrid models, implicit modeling, and direct dense landmark regression methods; see more details in Section~\ref{sec:relatedWork}.
However, regardless of the model type used, it is of great importance to have two intertwined, rather than decoupled, models for the geometry and texture, for more realistic results.
This is more crucial for synthesis tasks, where the newly generated faces must adhere to the combined geometry and texture manifold.
\begin{figure}[t!]
\centering
\includegraphics[width=0.95\linewidth]{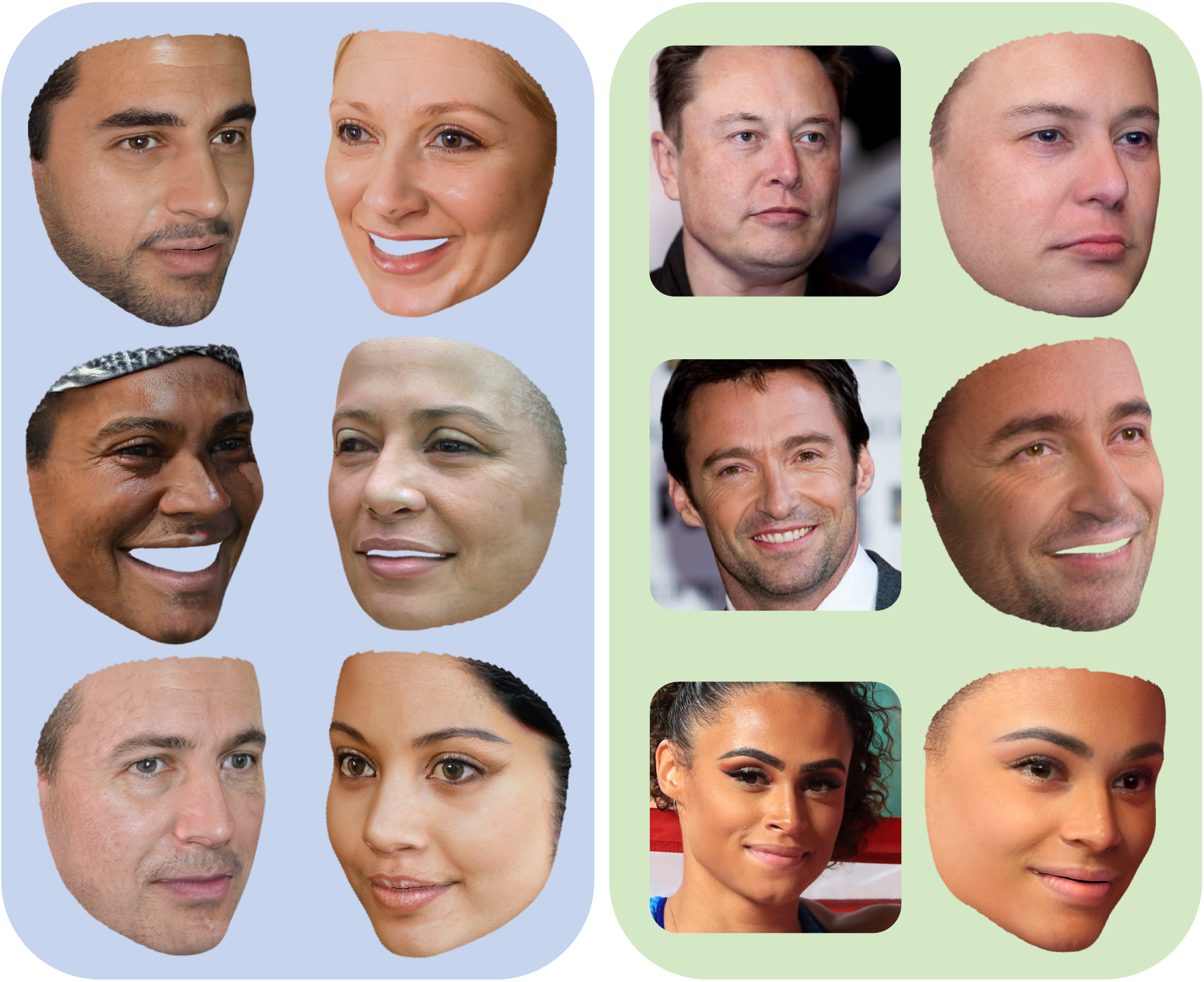}
\captionof{figure}{\textbf{Left:} Generation. \textbf{Right:} Reconstruction.}
\label{fig:teaser}
\end{figure}

In previous efforts, training a synthesis model for facial geometry and texture was either: (i) dependent on 3D facial scans (\ie via supervised learning) and produces high-quality results, or (ii) dependent on 2D facial images only (\ie unsupervised or semi-supervised), but produced low-quality, inaccurate, or incomplete models; see detailed overview in Section~\ref{sec:relatedWork}. 
Our work combines the best of both worlds, and provides an unsupervised training pipeline, independent of 3D facial scans, producing state-of-the-art modeling results on par even with fully supervised methods. 
Our high-resolution model is achieved by incorporating a linear as well as a direct regression facial model, a pre-trained 2D generative model, a deep feature manipulation component, and a differentiable rendering layer, as the building blocks for our unsupervised training pipeline. 

\section{Background and Related Efforts} \label{sec:relatedWork}

Next, we review related efforts.
Techniques that have been incorporated in the proposed pipeline are described in detail.

\paragraph{The 3D Morhpable Model (3DMM)~\cite{blanz1999morphable}} is arguably the most commonly used model both when generating or reconstructing facial geometries and textures; see survey~\cite{egger20203d} and Section~\ref{sec:relatedWork}.
The 3DMM model is obtained by semantically aligning facial scans to a template model comprised of $n$ vertices and performing PCA analysis~\cite{jolliffe1986principal} on the geometry, texture and expression vectors. The obtained $k$ principal components for shape and expression $\mathbf{U}\in\REAL ^{3n\times k}$ and mean shape $\mathbf{M}\in\REAL^{3n}$ comprise the 3DMM model. Given a set of shape and expression parameters $(\mathbf{p}_s\in\REAL^{k},\mathbf{p}_e\in\REAL^{k})$ the facial geometry is constructed as $\mathbf{S}=\mathbf{M}+ \mathbf{U}_s\cdot \mathbf{p}_s+\mathbf{U}_e\cdot \mathbf{p}_e$.
Texture modeling and formation are produced per-vertex in a similar manner.
Many improvements were suggested, for example,  ~\cite{booth20163d,booth2018large}, who improve the data acquisition and registration processes.
However, due to their linear nature, such models usually produce unrealistic over-smoothed samples~\cite{slossberg2018high}. 
As a remedy, many non-linear models have shown successful results in recent years; see more details in what follows.


\paragraph{3DMM fitting.} Given a 2D face image and the 3DMM geometry and expression basis, the goal of \emph{3DMM fitting (or, regression)} is to recover the 3DMM geometry and expression coefficients as well as a 3D rigid transformation $(\mathbf{R},\mathbf{t})$.
This is achieved by minimizing some rendering loss between the given 2D image and a face rendered using the above coefficients and parameters. Numerous approaches have been suggested for tackling this problem, ranging from optimization-based methods, like ~\cite{blanz1999morphable, gecer2019ganfit}, to one-shot deep learning pipelines such as the pioneering papers by~\cite{richardson20163d,zhu2016face}, more recently followed by~\cite{tewari2017mofa,genova2018unsupervised,deng2019accurate, guo2020towards} to name a few.
In our proposed texture generation pipeline we utilize the work of~\cite{deng2019accurate}, due to their state-of-the-art precision in estimating the geometry, expression, texture, and illumination parameters, as well as their available code and pre-trained model; see Section~\ref{sec:method}.

\paragraph{Non-linear and hybrid model fitting.} 
Recent efforts have built upon classical 3DMMs, proposing both hybrid~\cite{richardson2017learning,tewari2017self,sela2017unrestricted, slossberg2018high, tewari2019fml,  gecer2019ganfit, chen2019photo, shamai2019synthesizing} and completely non-linear models~\cite{nonlinear-3d-face-morphable-model, yenamandra2021i3dmm}. These deep network-based methods may incorporate linear parts supplemented by DNN-based parts and in other cases are entirely network-based.
Some models are presented only in the context of monocular geometry and texture recovery while others are also utilized in the context of synthesis. 
We find that models dedicated to one of those tasks often do not extend well to the other.


\paragraph{Dense landmark regression.} In~\cite{kartynnik2019real}, a regression network is trained to predict a dense collection of landmarks directly on a given facial 2D image.
These landmarks represent the projected vertex locations of a 3D canonical facial model.
This method achieves better fitting to the facial image and is not constrained to the limitations of the linear model. 
On the other hand, the extracted facial geometry is less accurate and detailed. 
Hence, this method is better suited for AR and other image-related tasks, \eg mouth-region mask extraction from 2D images; see Section~\ref{sec:mouthMasking}.

\paragraph{Realistic 2D face generation.} 
In a long line of work culminating in \cite{karras2020training}, various models have been proposed for the task of 2D face image generation. Such models are capable of generating highly realistic 2D facial images as well as project real 2D facial images onto the model's latent manifold.
As our model aims to mitigate the need for 3D scans of facial textures, we heavily rely on well-established 2D facial generative models as the basis for our proposed pipeline, successfully harnessing their high level of realism.
Throughout the proposed pipeline, we utilize the architecture, training methodologies as well as pre-trained model weights produced by the seminal papers of Karras \etal~\cite{karras2019style, karras2020training}, which have been shown to produce high-resolution realistic images; see Section~\ref{sec:method}. 

\paragraph{Manipulating facial properties via deep feature mapping.}
Most synthesis methods described above, specifically~\cite{karras2019style}, learn to map an input random noise vector, through some latent representation, into a realistic 2D facial image.
Following this popular approach, a variety of papers have emerged which learn to manipulate this intermediate latent vector to change some desired facial properties in the output 2D image. 
Such manipulation can be either statistics-based~\cite{chen2018facelet} or, more often, learning based~\cite{tewari2020stylerig}; see survey in~\cite{tolosana2020deepfakes} and references therein.

As will be discussed in detail in Section~\ref{sec:method}, our pipeline makes use of a 2D facial image generator to compensate for the lack of 3D facial scans. 
However, it is very challenging to compensate for 3D geometry and full facial texture using only \emph{randomly generated} 2D facial images.
To this end, we utilize a component that can control the pose of those generated images and show that the \emph{controlled} 2D images indeed suffice for full-texture learning.
To this end, we utilize the work of~\cite{tewari2020stylerig} for deep feature manipulation; see Section~\ref{sec:method}.

\begin{figure*}
    \centering
    \includegraphics[width=\textwidth]{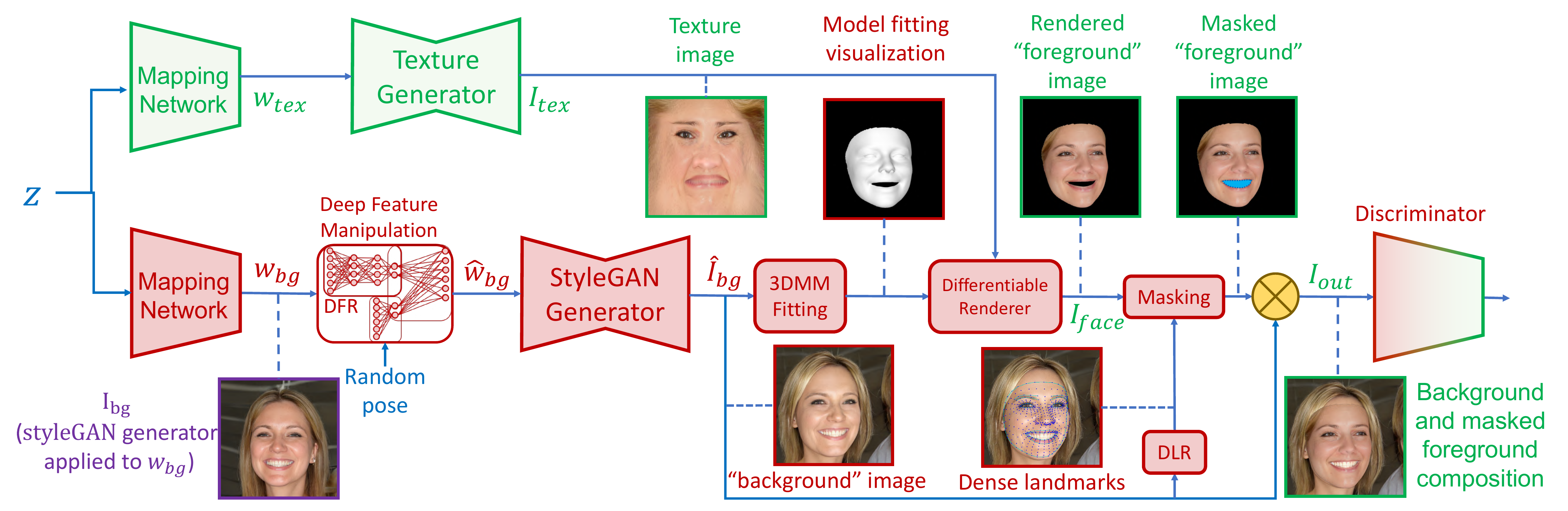}
    \caption{\textbf{Our training pipeline. }A vector $\mathbf{z}\in \REAL^{512}$ of Gaussian random noise is plugged into two mapping networks~\cite{karras2020analyzing} with the same architecture, producing two latent vectors $\mathbf{w}_{tex}, \mathbf{w}_{bg} \in \REAL^{18\times512}$, respectively. 
    The facial image encoded within $\mathbf{w}_{bg}$ is illustrated in purple.
    The vector $\mathbf{w}_{bg}$ is plugged into a deep feature manipulation network~\cite{tewari2020stylerig} called \stylerig{} to obtain the (manipulated) latent vector $\mathbf{\hat{w}}_{bg}\in \REAL^{18\times512}$ which encodes the same facial information as $\mathbf{w}_{bg}$ (\eg facial expression, identity, lighting, etc.) but with a modified facial orientation or expression. 
    We then feed $\mathbf{w}_{tex}$ and $\mathbf{\hat{w}}_{bg}$ into our texture and pre-trained \stylegan{} generators $G_T$ and $G_b$, outputting a texture image $I_{tex}$ and a 2D facial image $\hat{I}_{bg}$ respectively. We then recover 3DMM parameters $\hat{\mathbf{p}}_{bg}$ that best fit $\hat{I}_{bg}$~\cite{deng2019accurate} (illustrated by the textureless facial geometry), and use them to render the texture $I_{tex}$ into a 2D face image $I_{face}$ superimposed on $\hat{I}_{bg}$. We then mask out the mouth area of $I_{face}$ using a mouth mask (in bright blue) recovered from $I_{bg}$ using a dense landmark regression (DLR) model~\cite{kartynnik2019real}.
    The masked facial (foreground) image and the (background) image $\hat{I}_{bg}$ are then composed together to form the output image $I_{out}$. Finally, $I_{out}$ is fed into a pre-trained discriminator which is further trained. Trainable and pre-trained models are depicted in green and red respectively.}
    \label{fig:pipeline}
\end{figure*}

\paragraph{Reconstruction. }
In a long line of research, many methods have been suggested for 3D face reconstruction from a given 2D image. 
In~\cite{richardson20163d, richardson2017learning, sela2017unrestricted, dou2017end}, a mapping from 2D images to a 3D geometric representation is learned based on synthetic data pairs. 
Real facial textures were used, for example, in~\cite{gecer2019ganfit, gecer2021fast}, to obtain, in a supervised manner, a realistic reconstruction. 
However, acquiring such textures requires 3D scanning of faces, which is very costly and laborious, hence, impractical to scale to large numbers.
In this paper, we provide an unsupervised alternative that simultaneously requires no 3D scans, and achieves either comparable or higher quality reconstructions.
A pipeline for completion of a facial texture containing large holes was suggested in~\cite{deng2018uv}. A different approach that learns to ignore incomplete regions via masking was suggested in~\cite{shamai2019synthesizing}.
However, those also rely on textures as training data. 
A one-shot learning approach was proposed in~\cite{gecer2021ostec} which applies an iterative and very slow optimization process to complete a facial texture. 
The only other unsupervised reconstruction model we know of was proposed by~\cite{lin2020towards}. This work utilizes Graph Convolutional Networks to generate impressive high-quality reconstruction results; However, by not basing their pipeline on a 2D image generator which can produce controlled 2D images (\eg \stylegan{} combined with a model as~\cite{tewari2020stylerig}), their method is, by design, not intended for the task of generation of expressive 3D models, might lack important facial details, and does not account for the coupling of geometry and texture.  See detailed comparison with ~\cite{lin2020towards} in Section~\ref{sec:exp_results}.

\paragraph{Generation. }
While most prior efforts have focused on reconstruction, some methods have been proposed for the generation of random but realistic facial models. 
In ~\cite{marriott20213d}, who focus on improving facial recognition models via synthetic augmentation a GAN-based approach is proposed as well; however, their pipeline focuses more on controlling model parameters intending to supplement the training data for facial recognition models. This focus, as opposed to the realistic generation of completely random faces, leads to a less desirable outcome in terms of realism and resolution.
Hence, the results are not visually pleasing as depicted in Fig.~\ref{fig:gen_results}. 
In~\cite{slossberg2018high, shamai2019synthesizing, gecer2019ganfit}, 3DMMs combined with generative models were used for either generation or reconstruction of realistic textures. 
However, these methods use high-quality facial scans during training, which were obtained by specialized facial scanners and are not publicly available. 
This makes these methods very difficult to utilize in practice. 
Another downside is that limited scanned data is far less diverse than abundant facial 2D images commonly found in many datasets.

The following question now naturally arises: \textbf{can we generate or reconstruct high quality, realistic, and diverse 3D facial texture and corresponding geometry, by learning from 2D image data only?}

\subsection{Our contribution}
The main contributions of our method are the following:

\textbf{(i)} We affirmatively answer the question above and provide the first unsupervised high-fidelity generation as well as reconstruction pipeline capable of producing realistic textures coupled with corresponding geometries. 

    
\textbf{(ii)} Our pipeline decouples intrinsic texture features related to the person's identity, from extrinsic properties such as pose and lighting. This allows us to tackle the challenging problems of reconstructing a facial texture including full high-quality side views, as well as performing model re-illumination. See Section~\ref{sec:stylerig}.

\textbf{(iii)}
We present state-of-the-art results in both model generation as well as full texture recovery. We support this claim via both qualitative as well as quantitative results and comparisons. See Section~\ref{sec:exp_results}.

\textbf{(iv)} Our results are fully reproducible as only freely available datasets and models are required during training and inference. In addition, we provide all our trained model weights for both generation and reconstruction tasks~\cite{pretrainedModels}.

\section{Unsupervised Learning of Facial Textures and Geometries} \label{sec:method}
In this section, we dive into the details of our unsupervised 3D facial geometry and full texture generation pipeline and its components. 
While our pipeline learns to recover both geometry and texture, our novelty lies mainly in high-quality texture generation and retrieval, as we believe that the main effect on the perception of model realism stems from high-resolution texture rather than highly detailed geometry. This was also noted by~\cite{slossberg2018high} who demonstrated this point by varying texture and geometry quality while observing the overall effect on model realism. 
Nevertheless, recovery of highly detailed geometry is still an important research topic with many successful efforts such as ~\cite{richardson2017learning,sela2017unrestricted,tran2018extreme,chen2019photo} to name a few.

\begin{figure*}
    \centering
    \includegraphics[width=\textwidth]{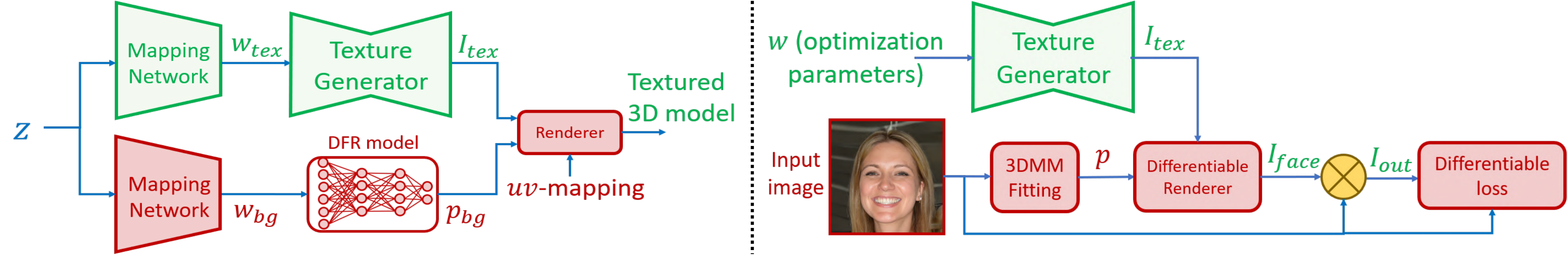}
    \caption{\textbf{Our inference pipelines.} During inference, we drop some components related to the training pipeline (see Fig.~\ref{fig:pipeline}). 
    \textbf{(Left) Generation: }
    As before, a single latent vector $\mathbf{z}$ is used to generate $\mathbf{w}_{tex}$ and $\mathbf{w}_{bg}$ via two mapping networks.
    The latent vector $\mathbf{w}_{bg}$ is used to generate 3DMM geometry parameters $\mathbf{p}_{bg}$ via the trained DFR model while $\mathbf{w}_{tex}$ is introduced to the trained texture generator yielding the corresponding texture image $I_{tex}$. 
    The parameters $\mathbf{p}_{bg}$ are used, along with our canonical UV parametrization, to generate the 3DMM geometry which we render using $I_{tex}$ as the mesh texture.
    \textbf{(Right) Reconstruction: }A given input image $I$ is first plugged into a fitting model producing its 3DMM parameters $\mathbf{p}$.
    A latent vector $\mathbf{w}$ containing our optimization parameters is then inserted into our trained texture generator producing a texture image $I_{tex}$. Using a differentiable renderer, $\mathbf{p}$ and $I_{tex}$ are rendered into a 2D face image $I_{face}$, which is superimposed on $I$ to produce our output $I_{out}$. Finally, a VGG loss similar to~\cite{karras2020training} is evaluated between $I$ and $I_{out}$.}
    \label{fig:inference}
\end{figure*}

An overview of our training and inference pipelines is depicted in Figs.~\ref{fig:pipeline} and~\ref{fig:inference} respectively.
Our approach to unsupervised learning of facial textures utilizes an adversarial loss to train a texture generator, $G_T$, while harnessing a pre-trained 2D facial image generator, $G_{bg}$, in the following fashion.
We start by generating, via $G_{bg}$, a 2D facial image which we term a \emph{background image}. We then fit a corresponding geometry to the background image using a pre-trained 3DMM fitting model \cite{deng2019accurate}. We then proceed to generate a facial texture $I_{tex}$ via our trainable texture generator $G_{T}$. Our generated synthetic texture is stored as a 2D image coupled with a canonical UV parametrization relating between image locations to the vertices of the 3D facial model. The model fitted to the background image enables the seamless mapping of our synthetic texture image $I_{tex}$ into the background image as depicted in Fig.~\ref{fig:pipeline}.

Our texture generator is trained within a GAN framework for which a discriminator model, serving as a trainable loss function, is trained to differentiate between blended and real images and thus continuously improves the generator quality. 
In order to generate high-resolution facial textures from all viewing angles, it is crucial to control various properties within the images generated by $G_{bg}$.
For example, we require that each generated identity appears under a range of poses. We, therefore, utilize a deep feature manipulation component, as proposed by \cite{tewari2020stylerig}, that encodes the desired properties within the input of $G_{bg}$. In addition, in order to disentangle between the albedo and shading components of the texture, we estimate the lighting conditions in $G_{bg}$ and apply them to our texture within the rendering process. Section~\ref{sec:stylerig} further elaborates on these critical components.

\paragraph{Learning from 3D facial scans.} Prior efforts approached the task of training facial texture models by relying on difficult-to-obtain 3D scans. For example, in~\cite{slossberg2018high, gecer2019ganfit}, high resolution scans obtained by a 3DMD scanner are geometrically aligned and mapped to a canonical 2D domain. The mapped textures are used as training data for a GAN which is tasked to generate new and realistic ones. This methodology suffers from several drawbacks, \eg, (i) The 3D scans are not easily obtained or freely distributed, thus posing a significant barrier in reproducing such models. (ii) High-quality 3D scanners are expensive and cumbersome, limiting the ability to collect data. Hence, even when available, such datasets are comprised of at most a few thousand subjects, which can not encompass the huge variety of human faces. We mitigate the above issues by eliminating the dependency on scans and replacing them with widely available 2D facial images, thus providing a more accessible method and producing a more diverse texture model.

\paragraph{Obtaining 3D facial models from 2D images.} To replace the difficult-to-obtain 3D facial scans with prevalent 2D facial images, it is common to utilize a differentiable rendering layer. The rendering of 3D textured models into 2D images enables the utilization of 2D image-related architectures and losses.
This process also requires a 3D mesh, usually represented by a pair $(V,Tri)$ of vertex coordinates $V \in \REAL^{N_v\times 3}$ and triangulation $Tri \in \REAL^{N_f\times 3}$, as well as a $uv$ parametrization $\phi: \{1,\cdots,N_v \} \to [0,1]\times[0,1]$ that maps every vertex to coordinates on the canonical plane.
The vertex coordinates are first projected onto the 2D camera plane and the final pixel colors are determined by a rasterization process mapping the facial texture onto the projected mesh according to the predetermined UV parametrization, to obtain the desired facial rendering $I_{face}$.

Using this methodology, we can transform our training losses from the 3D to the 2D facial domain. We can thus incorporate the vast corpus of prior art regarding 2D images, including pre-trained models as well as large, high resolution, and freely available datasets; see Section~\ref{sec:transferLearning}.

Having established the above, the question remains how to obtain synthetic facial renderings which are indistinguishable from real facial images, considering that the rendered images lack hair, ears, inside of the mouth, background, etc.
Possible solutions include: \textbf{(i)} To segment the foreground (the face) in the real images, and ignore the rest of the image. Thus both real and rendered images contain a foreground (facial) region, imposed on an empty background.
Since the segmentation of the real images is not correlated with the geometry which produces the fake renderings, the resulting foreground can be easily distinguished from the rendered examples. 
\textbf{(ii)} Alternatively, one can apply the same geometry fitting methodology used during the generation of the fake images, to the real facial images. This fitting provides the facial boundary, which can be used to mask out the background. While this yields better results, the geometry fitting is not perfectly aligned at the pixel level causing easily distinguishable artifacts at the face boundary.

To overcome the above limitations, we propose to generate an additional 2D facial image $I_{bg}$, \eg using \stylegan, and utilize $I_{bg}$ as the background to our (foreground) rendered facial image $I_{face}$.
This is achieved by first fitting a geometric model to $I_{bg}$, which serves as the 3D mesh required for rendering $I_{tex}$ into a 2D image $I_{face}$, as previously detailed. 
This process embeds our synthetic facial texture image $I_{tex}$ into $I_{bg}$, enforcing the facial texture to be generated in a way that realistically blends with the surrounding parts in $I_{bg}$ (\eg, hair and ears); see Fig.~\ref{fig:pipeline}.

\subsection{Transfer learning} \label{sec:transferLearning}
The process described above results in a 2D facial image, enabling the use of standard 2D image losses. 
As common in generative models, we use an adversarial loss to discriminate between real and fake images. 
Fortunately, many such pre-trained GANs are available for the task of 2D facial image generation~\cite{karras2020training}. 
For the mapping network, texture generator, and discriminator, we use the architecture proposed in \stylegan2~\cite{karras2020analyzing}.
As facial textures are closely related to 2D facial images, we initialize the above models with the pre-trained \stylegan2 weights.
This transfer learning approach has dramatically reduced our pipeline training time and improves texture quality, as was also reported by~\cite{karras2020training}.

\subsection{Pose and illumination invariant textures} \label{sec:stylerig}
As detailed above, our unsupervised approach relies on rendering 2D images from the generated textures. 
However, this approach, without further improvements, has one inherent problem: when a vast majority of the background images $I_{bg}$ contain, for example, frontal faces, our textures generator adequately learns to generate textures with high-resolution frontal face details but fails to produce high-resolution details on the periphery of the face. 
We propose to solve this issue by introducing random facial rotations during training via deep feature manipulation.

In addition, without properly addressing scene lighting, the generator will incorporate the lighting into the generated texture; see Fig.~\ref{fig:ablation}. 
However, it is desirable to decouple the albedo from the illumination effects, enabling post-relighting of the texture. Hence, we relight the models during training using the lighting parameters recovered by~\cite{deng2019accurate}.

\paragraph{Deep feature pose manipulation. }
In order to overcome the first problem above, we manipulate the latent vector $w_{bg}$, from which the background image $I_{bg}$ is generated, to enforce the generation of faces in a variety of orientations.
To this end, we perform deep feature manipulation by adopting the methodology proposed in~\cite{tewari2020stylerig}. 

The manipulation model, termed \stylerig, is comprised of two parts. A Differentiable Face Reconstruction Network, or DFR model, which takes as input the latent vector $\mathbf{w}$ and produces estimated 3DMM parameters $\mathbf{p} = DFR(\mathbf{w})$ which include $(\mathbf{p_s},\mathbf{p_e},\mathbf{p_t},\bgamma,\mathbf{R}, \mathbf{t})$, shape, expression, texture and lighting, rotation and translation parameters respectively. We train our model utilizing the highly versatile 3DMM model generated by~\cite{booth20163d}

A second network termed \stylerig{} takes as input a latent vector $\mathbf{w}$ and a set of parameters $\mathbf{p}$ and outputs a modified set of latent parameters $\hat{\mathbf{w}}$, where ideally the image produced by $I=G_{StyleGan}(\hat{\mathbf{w}})$ portrays the face $G_{StyleGan}(\mathbf{w})$ produced by $\mathbf{w}$ but modified to fit the parameters $\mathbf{p}$. 
In order to produce a rotated version of $I_{bg}$ we first modify the rotation parameters of $\mathbf{p}_{bg} := DFR(\mathbf{w}_{bg})$ to derive $\hat{\mathbf{p}}_{bg}$ and then apply $\hat{\mathbf{w}}_{bg}=Stylerig(\hat{\mathbf{p}}_{bg},\mathbf{w}_{bg})$. The image $\hat{I}_{bg}$ derived from $\hat{\mathbf{w}}_{bg}$ contains a rotated version of the same person as in $I_{bg}$. 
We then generate a texture image using the latent vector $\mathbf{w}_{tex}$, regardless of the rotation angles which were modified in $\mathbf{w}_{bg}$. This yields the desired pose-invariance within the texture generator; see Fig.~\ref{fig:pipeline} 

The same DFR model used above will be also utilized during inference in order to recover corresponding geometries for our generated textures; see Section~\ref{sec:recoverGeom} and Fig.~\ref{fig:inference}. This allows us to efficiently generate corresponding geometries directly from latent vectors without using the trained \stylegan generator during inference.

\paragraph{Training for re-illumination.}
To generate textures with no illumination effects, we first estimate the background scene lighting and relight the texture during training. Assuming a Lambertian reflectance model, we estimate the parameters $\mathbf{\gamma}_b \in \REAL^{3x9}$ from $I_{bg}$, as coefficients of $9$ Spherical Harmonics (SH) basis functions~\cite{ramamoorthi2001efficient,ramamoorthi2001signal} for R,G and B illumination bands, and relight the rendered image $I_{face}$ under the recovered illumination. The coefficients $\mathbf{\gamma}_b$ with the computed vertex normals $\mathbf{n}_i$ and SH functions $\Phi_b$ produce the per-vertex lighting value as $\mathbf{C}(\mathbf{n}_i| \bgamma) = \sum_{b = 1}^{B^2} \gamma_b \Phi_b(\mathbf{n}_i)$. 
We perform two renderings, one for illumination and one for albedo, and perform pixel-wise multiplication to derive the illuminated rendering
\begin{equation*}
I_{face} = \mathcal{R}(\mathbf{S}(p_s,p_e),G(w_{tex})) \cdot \mathcal{R}(\mathbf{S}(p_s,p_e),\mathbf{C}(\mathbf{n}_i| \bgamma)),
\label{eq:lighting_render}
\end{equation*}
where $\mathcal{R}(G,T)$ signifies the rendering operator applied to a geometry $G$ and a texture $T$, $\mathbf{p}_s, \mathbf{p}_e$ are respectively shape and expression parameters recovered from $I_{bg}$ and $w_{tex}$ is the input latent vector for the texture generator.

This process results in the texture generator producing textures with no baked-in lighting effects, \ie, textures that incorporate the albedo only, so that the re-illuminated texture via $\bgamma$ would match the lighting present in the background image $I_{bg}$ and seem realistic to the discriminator.

\subsection{Recovering corresponding geometry} 
\label{sec:recoverGeom}
As detailed in Section~\ref{sec:stylerig}, a trained DFR model can recover the geometry parameters (such as the parameters obtained via a geometry fitting pipeline, as in Sections~\ref{sec:relatedWork}). However, we note that the DFR extracts those parameters directly from a latent vector $w$ and not from an image.
As already part of our pipeline, the latent vector $w_{bg}$ is plugged into the deep feature manipulation component, which contains the DFR model; see Fig.~\ref{fig:pipeline}. 
Hence, as a byproduct of this pipeline, we obtain geometry parameters designated for the facial image to be later rendered. 
However, the simplicity of the DFR model and the fact that it does not directly observe the 2D image causes imprecise fitting results. 
In order to retain realism of the composed $I_{out}$ during training, we must employ better fitting methodologies onto $I_{bg}$ such as \cite{deng2019accurate}. 
We will find a use for the DFR model when performing 3D synthesis as it relates between latent and 3D model parameters effectively, thus, encoding the desired coupling between texture and geometry; see Fig.~\ref{fig:inference}. 

\subsection{Masking mouth area} \label{sec:mouthMasking}
As fitting a 3DMM to a 2D image is a challenging task, minor fitting errors are inevitable. 
The facial region mask rendered using the recovered 3DMM parameters thus might not perfectly align with the face in the 2D image, causing the (empty) mouth region in the mask to misalign with the mouth in the background image. 
Hence, for the rendered texture in the mask region to realistically blend with the background image, the generator is falsely forced to generate undesired details, \eg teeth on top of the lips region; see example in Fig.~\ref{fig:ablation}.
To prevent this from happening, we incorporate the work of~\cite{kartynnik2019real}, which performs a dense facial landmark regression on the 2D image, to find the mouth region with high accuracy. 
We then remove this predicted mouth region from the facial mask obtained by the 3DMM fitting procedure. 
This forces the rendered mask to contain an ``empty region'' in the mouth region of the background image. 
By further masking the inner-mouth area we can eliminate the appearance of teeth within our generated textures, as illustrated in Fig.~\ref{fig:ablation}.

\subsection{Fully unsupervised training}
The proposed pipeline above generates full facial textures along with corresponding geometries, and, using a differentiable renderer, synthesizes a 2D facial image that is plugged into a discriminator model. 
Besides the synthetic 2D images above, real 2D facial images are also fed into the discriminator during training.
Such 2D real images are widespread and can be taken from any dataset of facial images, \eg \cite{karras2019style}. 
Moreover, we can utilize the (already in use) \stylegan{} generator to generate such 2D images during training, rather than use an existing dataset. 
As the common 2D image datasets are huge, replacing them with a pre-trained model can be very useful when either the storage memory is limited or communication time is crucial.

\section{Experimental results}
\label{sec:exp_results}
\begin{figure}
\centering
\includegraphics[width=\linewidth]{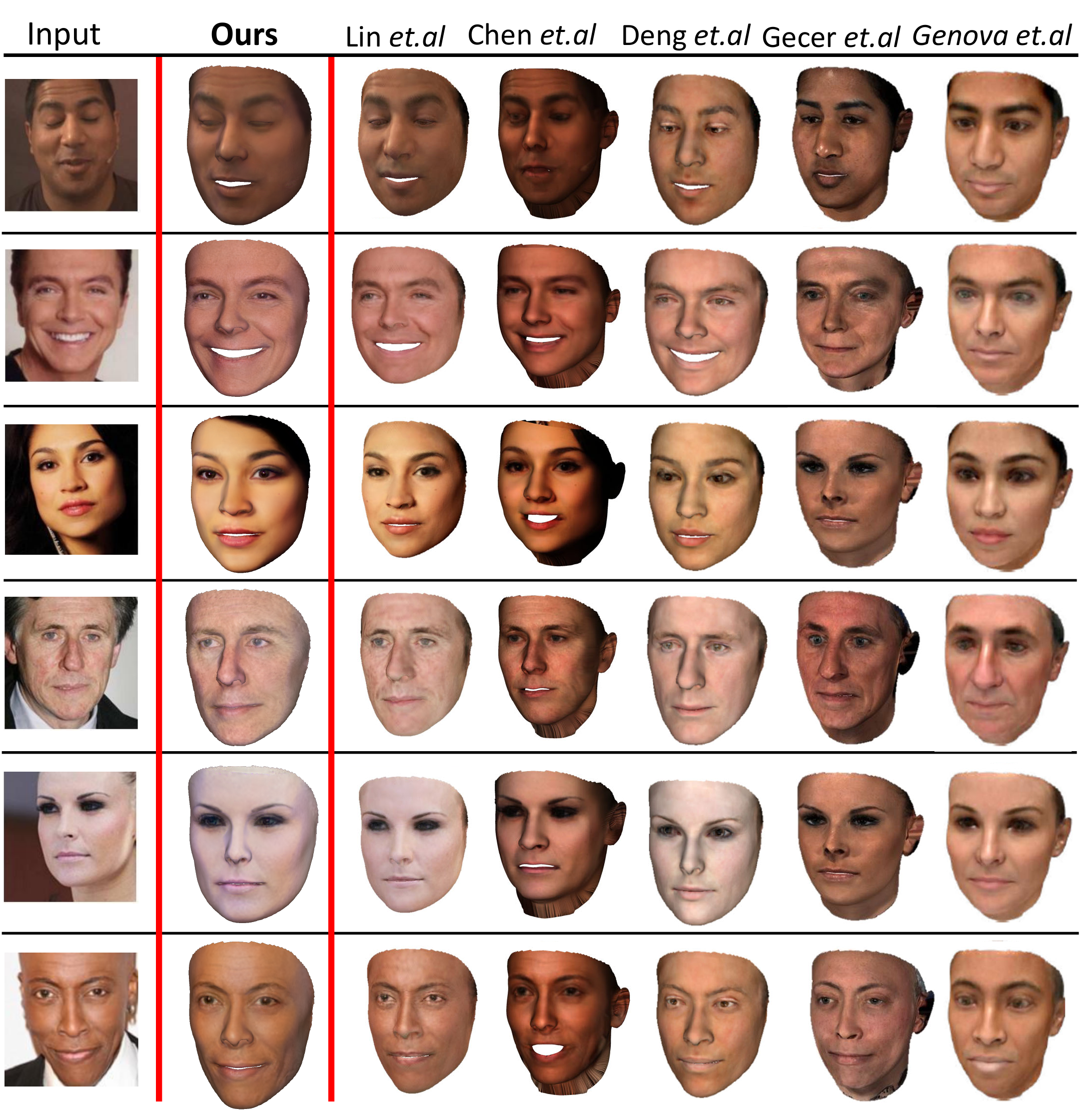}
\caption{\textbf{Qualitative Reconstruction Comparison Results:} We present our texture reconstruction results on the MOFA test-set~\cite{tewari2017mofa} and compare to previous results \cite{lin2020towards, chen2019photo, deng2019accurate, gecer2019ganfit, genova2018unsupervised}, respectively. This figure is best viewed when zoomed in.
}
\label{fig:qualitative}
\end{figure}
We compare our proposed approach to several state-of-the-art texture reconstruction and 3D generation methods, most of which utilize facial scans for training.
We provide quantitative as well as qualitative evidence demonstrating that our model outperforms previous methods, both supervised and unsupervised by scanned textures, in terms of texture reconstruction quality, realism, and details.

\textbf{Implementation details. }
We implemented our pipeline in Python using Pytorch~\cite{paszke2019pytorch} library and trained it on $4$ RTX 3090 GPUs on the FFHQ dataset~\cite{karras2019style}.
As mentioned in Section~\ref{sec:transferLearning}, we initialized our models from the pre-trained weights of \stylegan~\cite{karras2020analyzing}, using default parameters and their recommended losses.

\subsection{Face generation}
\begin{figure*}[t!]
\centering
\includegraphics[width=\linewidth]{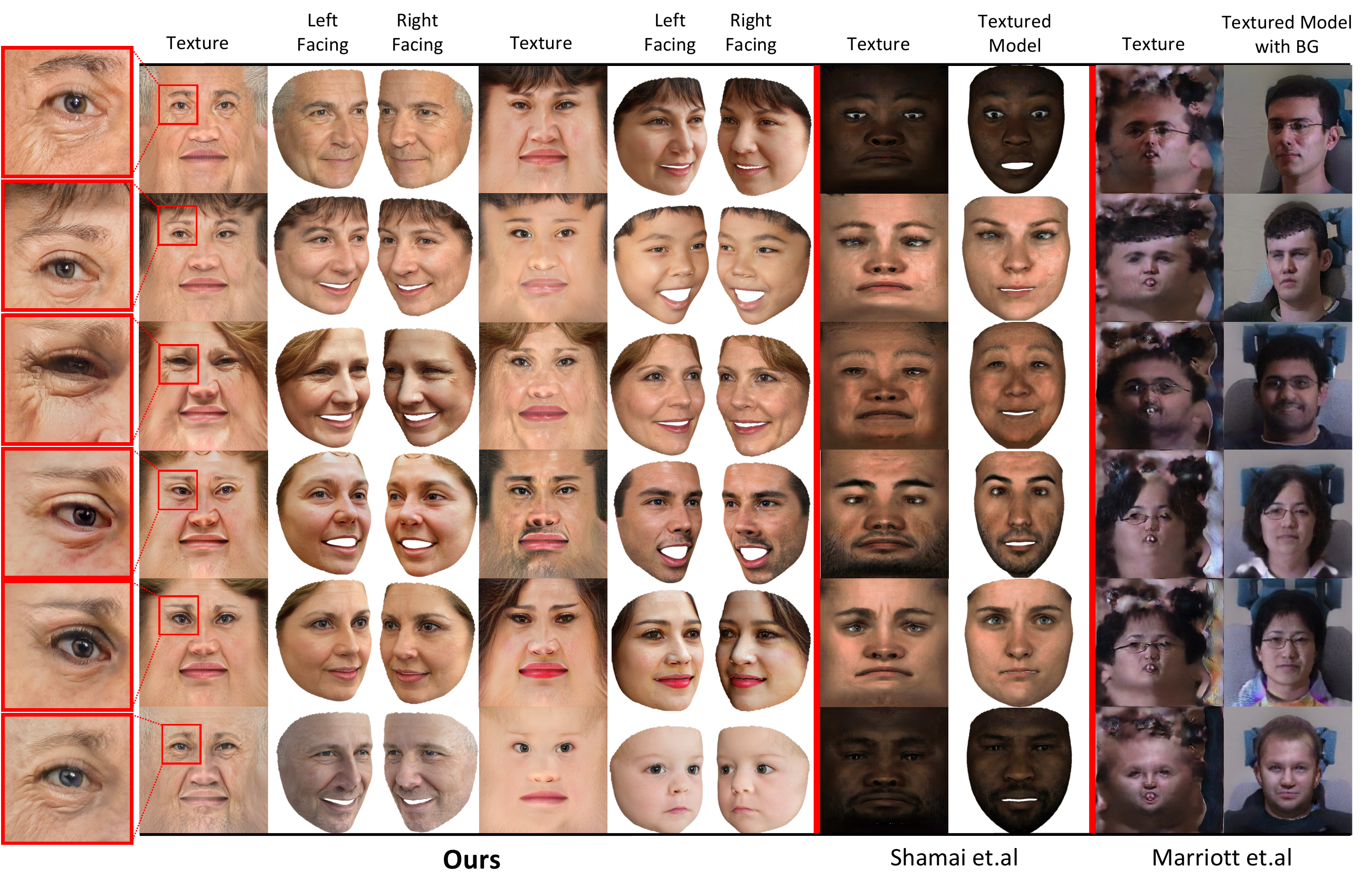}
\caption{\textbf{Facial Synthesis Comparison:} We visually compare our results both in texture and rendered textured geometries (in multiple views) to: Shamai \etal~\cite{shamai2019synthesizing} and Mariott \etal~\cite{marriott20213d}. Our high resolution textures provide highly realistic faces spanning a wide variety of ages, ethnicity and appearance. The leftmost column provides a zoomed-in crop, highlighting the high resolution realistic details. Our method still presents higher level detail and realism as compared to both previous methods, although~\cite{shamai2019synthesizing} is supervised by scanned textures.}
\label{fig:gen_results}
\vspace{-3mm}
\end{figure*}

We randomly generated textures and corresponding geometries via our inference pipeline; see Section~\ref{sec:method} and Fig.~\ref{fig:inference}. We present the texture images with zoomed-in areas to highlight the high level of detail and realism. We compare our results to the supervised model proposed by of~\cite{shamai2019synthesizing} as well as the unsupervised model from~\cite{marriott20213d}; see Fig.~\ref{fig:gen_results}. Additional randomly generated faces are included within the supplementary material.

\subsection{Facial texture reconstruction}
Fig.~\ref{fig:qualitative} presents a qualitative comparison between our texture reconstruction pipeline from Fig.~\ref{fig:inference} to several state-of-the-art prior ~\cite{lin2020towards, chen2019photo, deng2019accurate, gecer2019ganfit, genova2018unsupervised}. The comparison demonstrates that our model can reproduce challenging textures \eg difficult lighting conditions, makeup, and extreme expressions and compares favorably to previous approaches, including methods based on supervised training from 3D scans. Note that we utilize~\cite{deng2019accurate} for geometry recovery and thus focus our comparison on texture recovery only. Additional reconstruction results produced from high-resolution images are depicted in Fig.~\ref{fig:teaser} and the supplementary material. The results demonstrate that our model is capable of high-resolution texture recovery when presented with high-quality input images.

\subsection{Ablation study}
\begin{figure}
\centering
\includegraphics[width=0.85\linewidth]{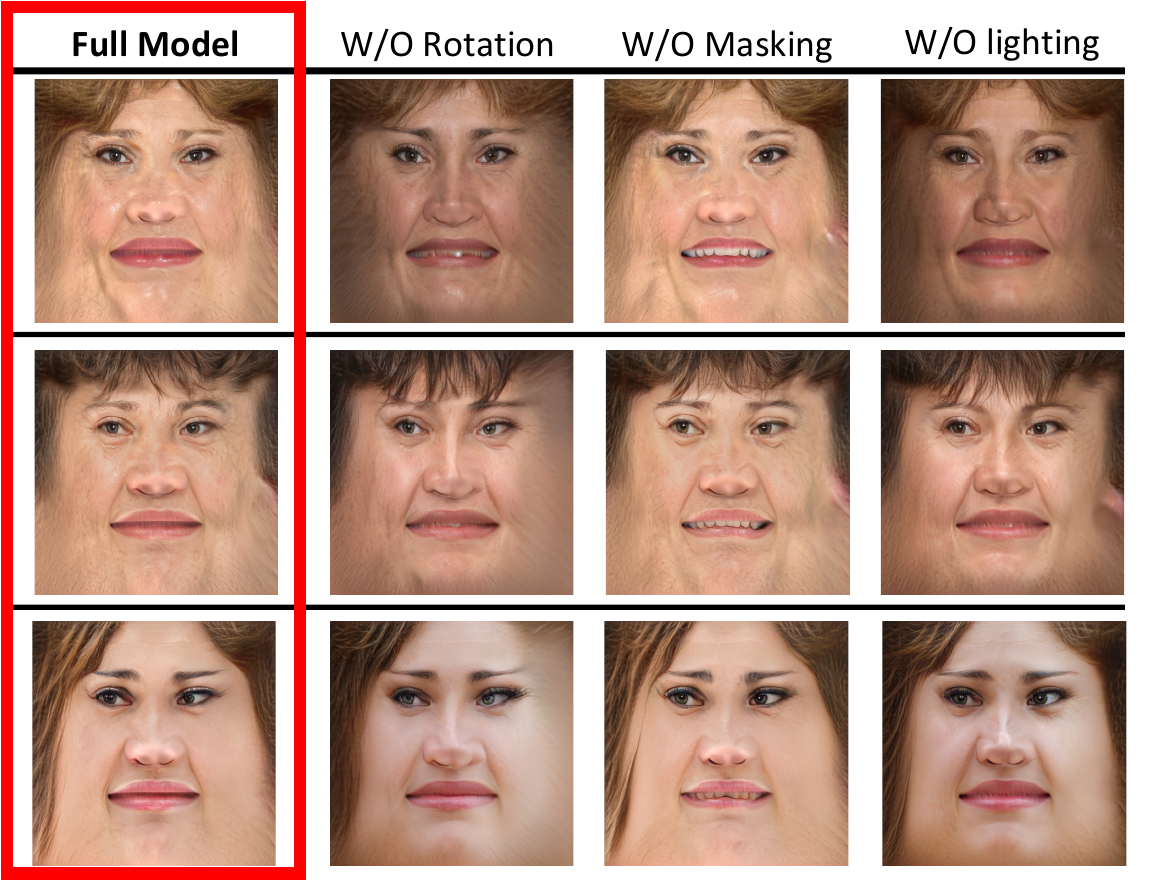}
\caption{\textbf{Ablation Study.} Left to right: full model, without rotations, without mouth masking, and without relighting. Removing those $3$ components yields poor details in the texture periphery, unwanted teeth, and baked-in lighting, respectively.
}
\vspace{-3mm}
\label{fig:ablation}
\end{figure}

In Fig.~\ref{fig:ablation} we present an ablation study, where the full proposed model is shown to consistently produce more realistic results compared to its variants with missing components. This suggests that each of our pipeline components is crucial for producing satisfactory output results. We show that: (i) model rotations during training are crucial for generating high details on the peripheral areas of the texture; see Section~\ref{sec:stylerig}, (ii) mouth masking eliminates the unwanted teeth artifacts; see Section~\ref{sec:mouthMasking}, and (iii) model illumination during training successfully disentangles albedo from shading, producing models that can be realistically integrated into scenes with varying lighting conditions; see Section~\ref{sec:stylerig}.

\subsection{Quantitative results}
\begin{table}

\begin{center}
\setlength{\tabcolsep}{3pt}{
\begin{tabular}{|c|cccc|}
\hline
Metric & \cite{deng2019accurate} &  \cite{marriott20213d} & \cite{deng2018uv} & \textbf{Ours} \\
\hline

$L_1$ distance $\downarrow$ & 0.052  & 0.034& / & \textbf{0.0244} \\
PSNR $\uparrow$ & 26.58  & 29.69 & 22.9$\sim$26.5 & \textbf{32.889}  \\
SSIM $\uparrow$ & 0.826 & 0.894 & 0.887$\sim$0.898 & \textbf{0.972} \\
LightCNN~\cite{wu2018light} $\uparrow$ & 0.724  & 0.900 & / & \textbf{0.96} \\
\hline

\end{tabular}}
\end{center}

\caption{\textbf{Quantitative Evaluation:} We evaluate our reprojected reconstruction similarity on the CelebA\cite{liu2015faceattributes} test-set. Our analysis shows that our method achieves better similarity scores in all reported metrics when compared to previous methods ~\cite{deng2018uv,deng2019accurate,marriott20213d}.}
\label{tab:quantitativ}
\vspace{-3mm}
\end{table}
Table~\ref{tab:quantitativ} presents our quantitative study on the task of texture reconstruction conducted on the CelebA~\cite{liu2015faceattributes} test-set, containing nearly 20k images. We test our reconstruction quality using three classical image reconstruction metrics and one based on a face recognition model~\cite{wu2018light}.
Our method achieves better scores in all tested metrics compared to the state-of-the-art methods~\cite{deng2019accurate,marriott20213d,deng2018uv}.
In contrast to \cite{marriott20213d}, we do not omit problematic areas by masking.

\section{Discussion, Limitations, and Future Work} \label{sec:discussion}
We introduced a novel unsupervised pipeline for both generation and reconstruction of high resolution realistic facial textures utilizing the acclaimed \stylegan{} framework. Our experiments demonstrate that we surpass prior art in both tasks, including models based on supervised training via scanned facial textures, in both reconstruction quality and generation realism. Our proposed model is capable of generating realistic geometries based on 3DMM modeling and importantly matches the geometry and texture via a single unified random input vector $\mathbf{z}$. We also plan to release our trained models including sample code for generation as well as reconstruction.

Due to the presence of subjects wearing glasses within the FFHQ dataset used for training, in some cases, our output texture might contain glasses. See Supplementary material for such examples.
This can be mitigated in future work by using latent feature manipulation or simply removing subjects with glasses from the training set.
In this paper, we did not utilize non-linear geometric representations as we note that high-resolution texture is the most crucial component in the quest for realistic facial generation. 
Improving geometry quality by utilizing non-linear representations and leveraging normal maps is left for future work.

{\small
\bibliographystyle{ieee_fullname}
\bibliography{references}
}


\clearpage
\appendix

\section{Additional Results}
\begin{figure*}

  \centering
\includegraphics[width=1\linewidth]{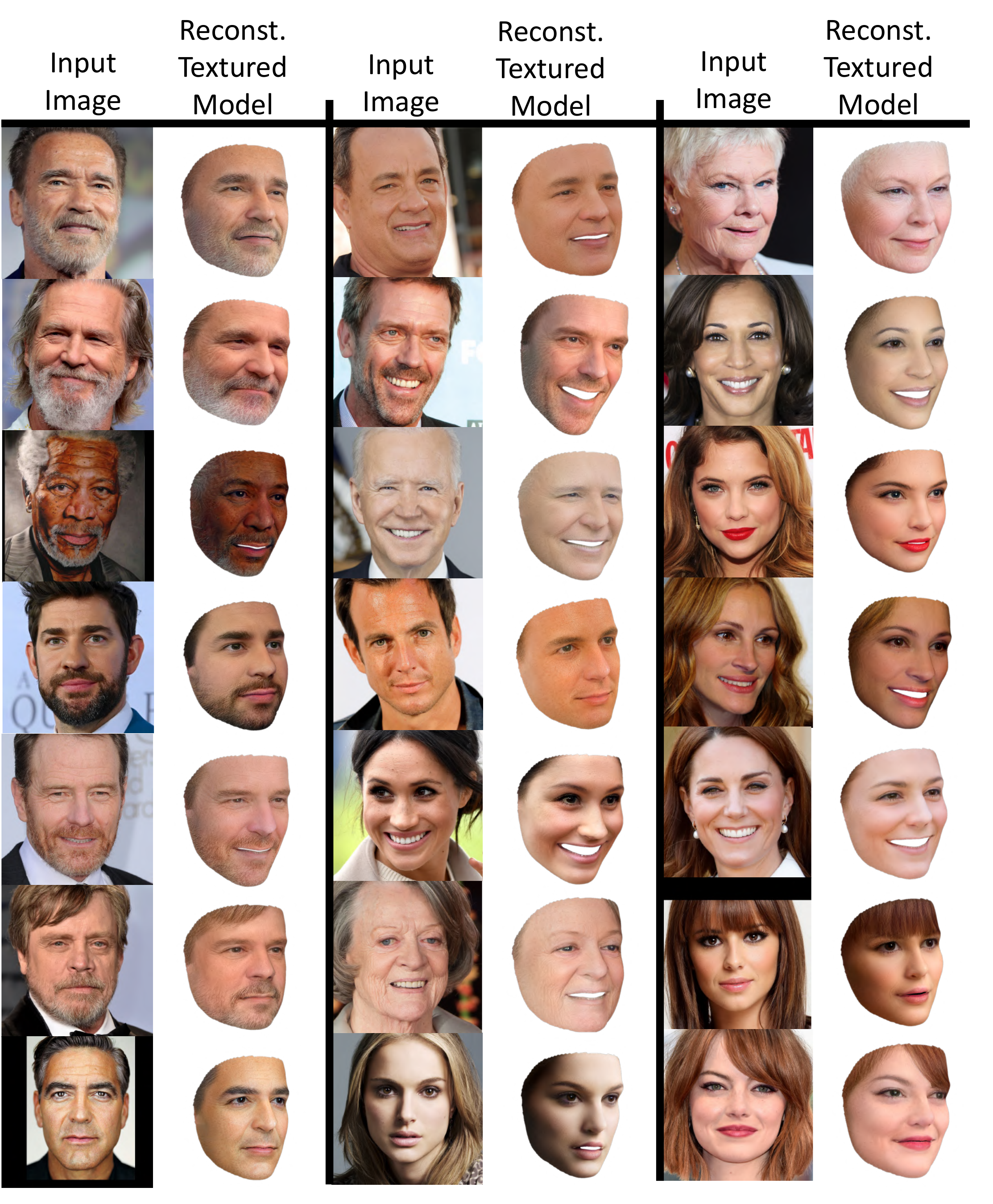}
\captionof{figure}{\textbf{Additional Qualitative Reconstruction Results.} We applied our reconstruction pipeline to numerous facial images. The 2D input images and the output textured models are presented side by side. The figure is best viewed when zoomed in.
}
\label{fig:recon_supp}
\end{figure*}

\begin{figure*}
\centering
\includegraphics[width=0.84\linewidth]{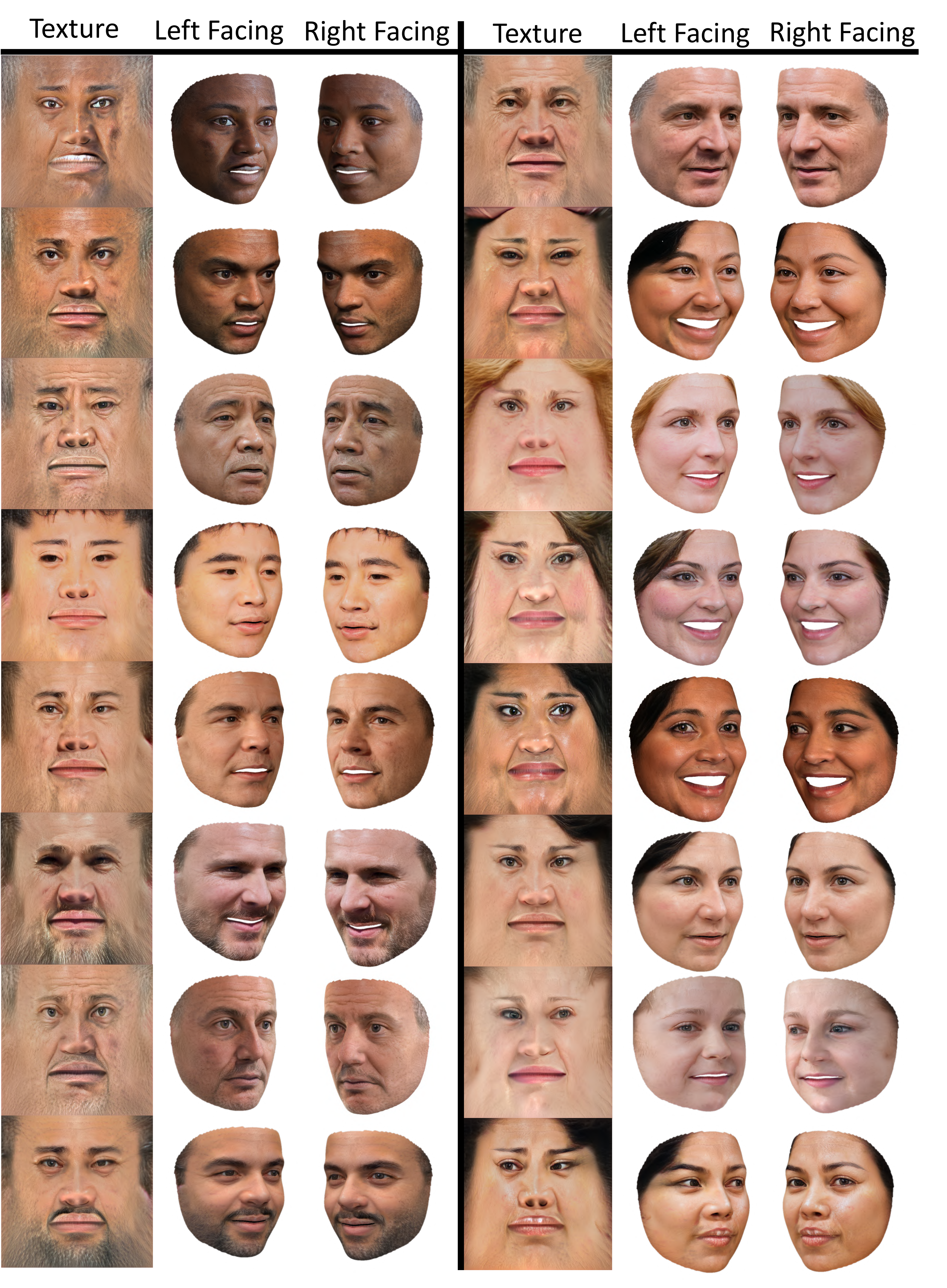}
\caption{\textbf{Additional Qualitative Generation Results.} We used our generation pipeline to generate various random textures and geometries. The figure is best viewed when zoomed in.
}
\label{fig:gen_supp}
\end{figure*}

We present additional results obtained by our method.
In Figure~\ref{fig:gen_glasses_supp} we demonstrated several generated results which include glasses. The glasses are generated within the facial texture causing a somewhat unrealistic result. This can be mitigated by latent feature manipulation or by simply eliminating samples with glasses from the training data. Figure~\ref{fig:recon_supp} demonstrates numerous reconstruction results obtained from our reconstruction pipeline. We observe highly detailed fully textured faces reconstructed from any input pose. In Figure~\ref{fig:gen_supp} we demonstrate more generation results, further demonstrating our high-fidelity texture generation capability coupled with matching realistic geometry. results are best viewed zoomed-in.
\begin{strip}
\centering
\includegraphics[width=0.45\textwidth]{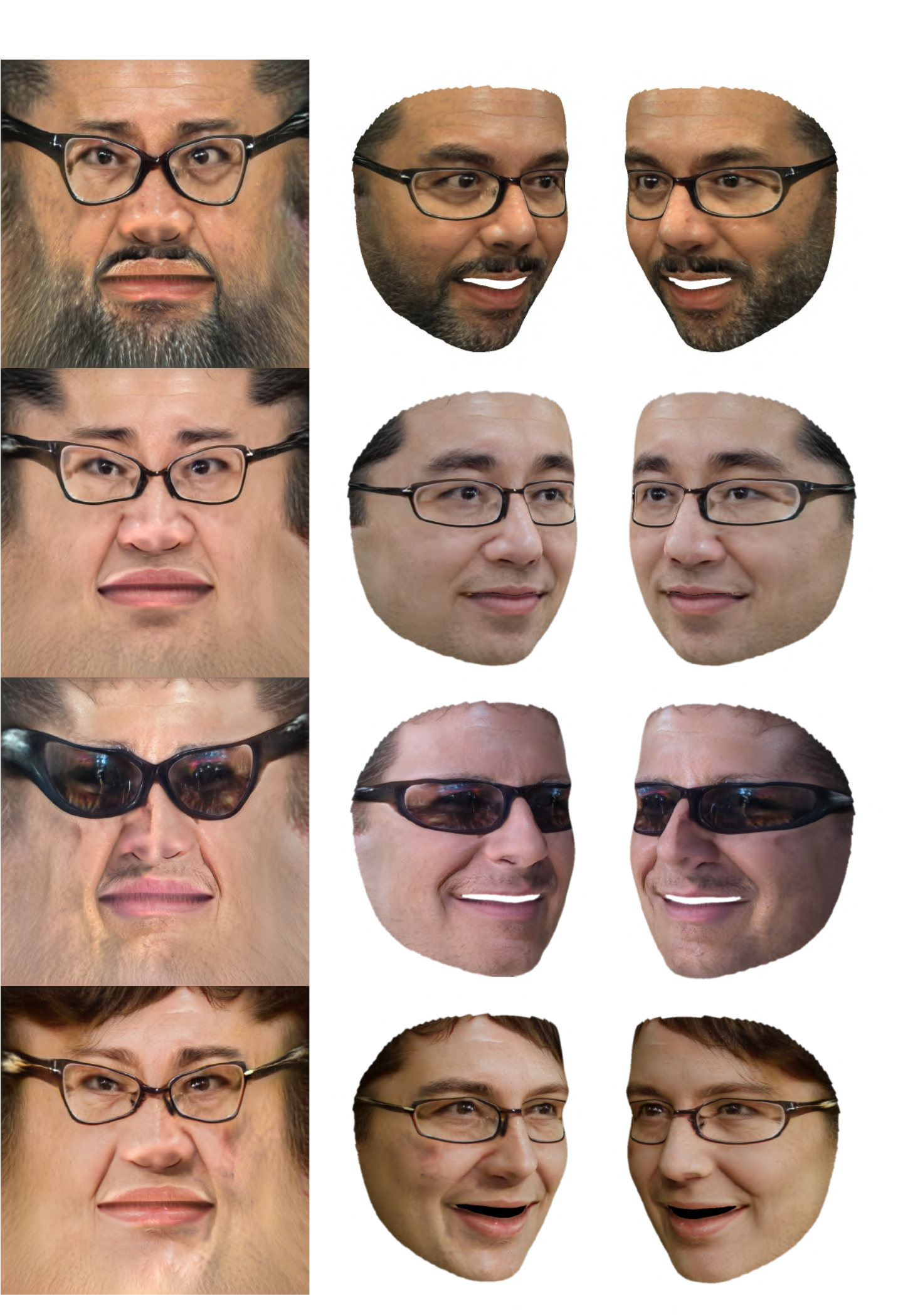}
\captionof{figure}{\textbf{Generation Results that Include Glasses.} Due to the presence of subjects wearing glasses within the FFHQ dataset used for training, in some cases, our output texture might contain glasses; see Discussion and Future Work Section.
}
\label{fig:gen_glasses_supp}
\end{strip}

\end{document}